\documentclass[final]{cvpr}

\usepackage{times}
\usepackage{epsfig}
\usepackage{graphicx}
\usepackage{amsmath}
\usepackage{amssymb}
\usepackage{xspace}
\usepackage{paralist}

\usepackage[pagebackref=true,breaklinks=true,colorlinks,bookmarks=false]{hyperref}

\def\cvprPaperID{1707} 
\def\confYear{CVPR 2021}

\begin{document}
\newcommand{\ChairSegments}[0]{Chair Segments\xspace}

\title{Chair Segments: A Compact Benchmark for the Study of Object Segmentation}

\newcommand*\samethanks[1][\value{footnote}]{\footnotemark[#1]}

\author{Leticia Pinto-Alva$^\ddagger$$^\dagger$, Ian K. Torres$^\natural$\thanks{Work conducted as visiting students at the University of Virginia}\,, Rosangel Garcia$^\S$\samethanks\,, Ziyan Yang$^\dagger$, Vicente Ordonez$^\dagger$\vspace{0.05in}\\
$^\ddagger$Universidad Católica San Pablo, $^\natural$University of Massachusetts, Amherst, $^\S$Le Moyne College, \\$^\dagger$University of Virginia\\
{\tt\small lp2rv@virginia.edu, zy3cx@virginia.edu, vicente@virginia.edu}
}

\maketitle

\begin{abstract}
    Over the years, datasets and benchmarks have had an outsized influence on the design of novel algorithms. In this paper, we introduce \ChairSegments, a novel and compact semi-synthetic dataset for object segmentation. We also show empirical findings in transfer learning that mirror recent findings for image classification. We particularly show that models that are fine-tuned from a pretrained set of weights lie in the same basin of the optimization landscape. \ChairSegments consists of a diverse set of prototypical images of chairs with transparent backgrounds composited into a diverse array of backgrounds. We aim for \ChairSegments to be the equivalent of the CIFAR-10 dataset but for quickly designing and iterating over novel model architectures for  segmentation. 
    On Chair Segments, a U-Net model can be trained to full convergence in only thirty minutes using a single GPU. 
   Finally, while this dataset is semi-synthetic, it can be a useful proxy for real data, leading to state-of-the-art accuracy on the Object Discovery dataset when used as a source of pretraining. Baselines and data can be found here: \url{https://github.com/uvavision/chair-segments}
\end{abstract}

\vspace{-0.2in}
\section{Introduction}
Many model architectures have been proposed for image classification with increasing levels of performance~\cite{krizhevsky2012imagenet,simonyan2014very,he2016deep,szegedy2014going,huang2017densely}. These advances have been fueled partially by the availability of common benchmarks that have allowed researchers to quickly iterate over model designs~\cite{lecun1998gradient,krizhevsky2009cifar,deng2009imagenet}. While the Imagenet dataset has brought significant advances, more compact datasets such as CIFAR-10 have allowed the quick exploration of novel ideas that are later validated in larger scale datasets. We aim to capture the simplicity of CIFAR-10 but for object segmentation. Current datasets for segmentation often require models that solve simultaneously object localization and object recognition on top of predicting segmentation masks (e.g.~Mask-RCNN~\cite{he2017mask}). We posit that as a result there is considerably less work in creating specialized segmentation networks that are solely designed to produce high quality segmentation masks (e.g.~U-Net~\cite{ronneberger2015u}). 

\begin{figure}[t]
\begin{center}
   \includegraphics[width=0.84\linewidth]{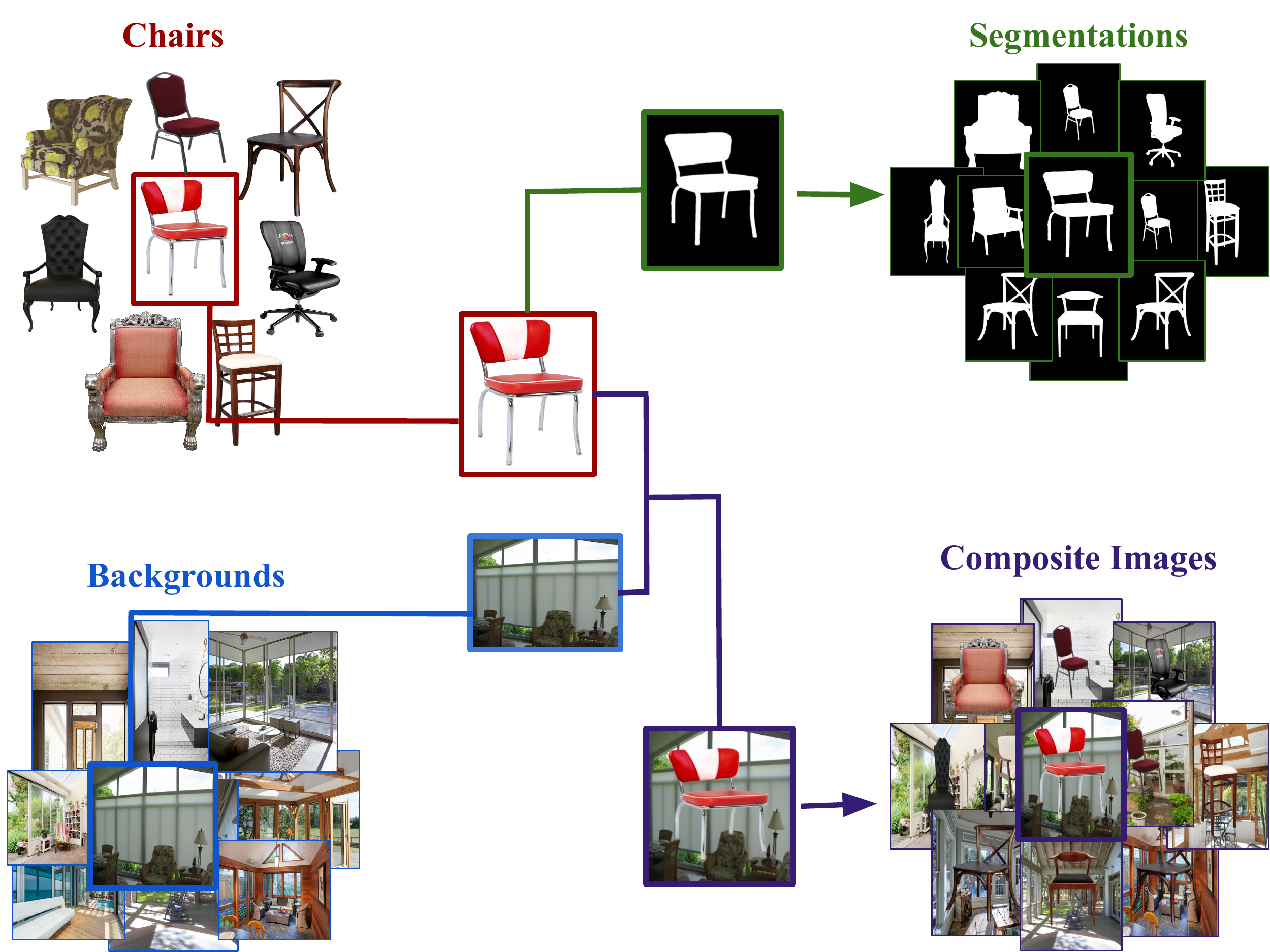}
\end{center}
   \caption{Images in the \ChairSegments dataset are created semi-synthetically by relying on images of chairs with transparent backgrounds and compositing them on background images to encourage diversity, and reasonably challenging testing conditions. Unlike other datasets, ground truth segmentation masks are pixel-level accurate as they are not hand annotated.}
\vspace{-0.2in}
\label{fig:lead-figure}
\end{figure}

In this work, we introduce Chair Segments, a semi-synthetic dataset that is compact, reasonably challenging, and has pixel-accurate ground truth segmentations (Figure~\ref{fig:lead-figure}). We focus on chairs, because this object is challenging both structurally and semantically. Chairs have also long interested perceptual psychologists and computer vision scientists because recognizing them requires cues that go beyond shapes and colors but also need to account for affordances~\cite{lloyd1997categorizing,lupyan2008chair,grabner2011makes,ordonez2014furniture,Polania_2020_CVPR_Workshops}, i.e. chairs are defined by their functionality.  In terms of structure chairs have elongated and complex shapes that could be hard to group. 
Moreover, we show that complex models such as U-Net can be reliably trained to full convergence in under an hour using a single GPU. We also demonstrate that these models are reliably modeling real-world data by showing state-of-the-art accuracy using transfer learning on the Object Discovery dataset (airplanes, horses, and cars)~\cite{Rubinstein13Unsupervised}. Finally, we use Chair Segments to replicate findings in the image classification domain where it was shown that models fine-tuned from a set of pretrained weights lie in the same basin of the optimization landscape~\cite{neyshabur2020being}.

\begin{figure}[t]
\begin{center}
   \includegraphics[width=0.88\linewidth]{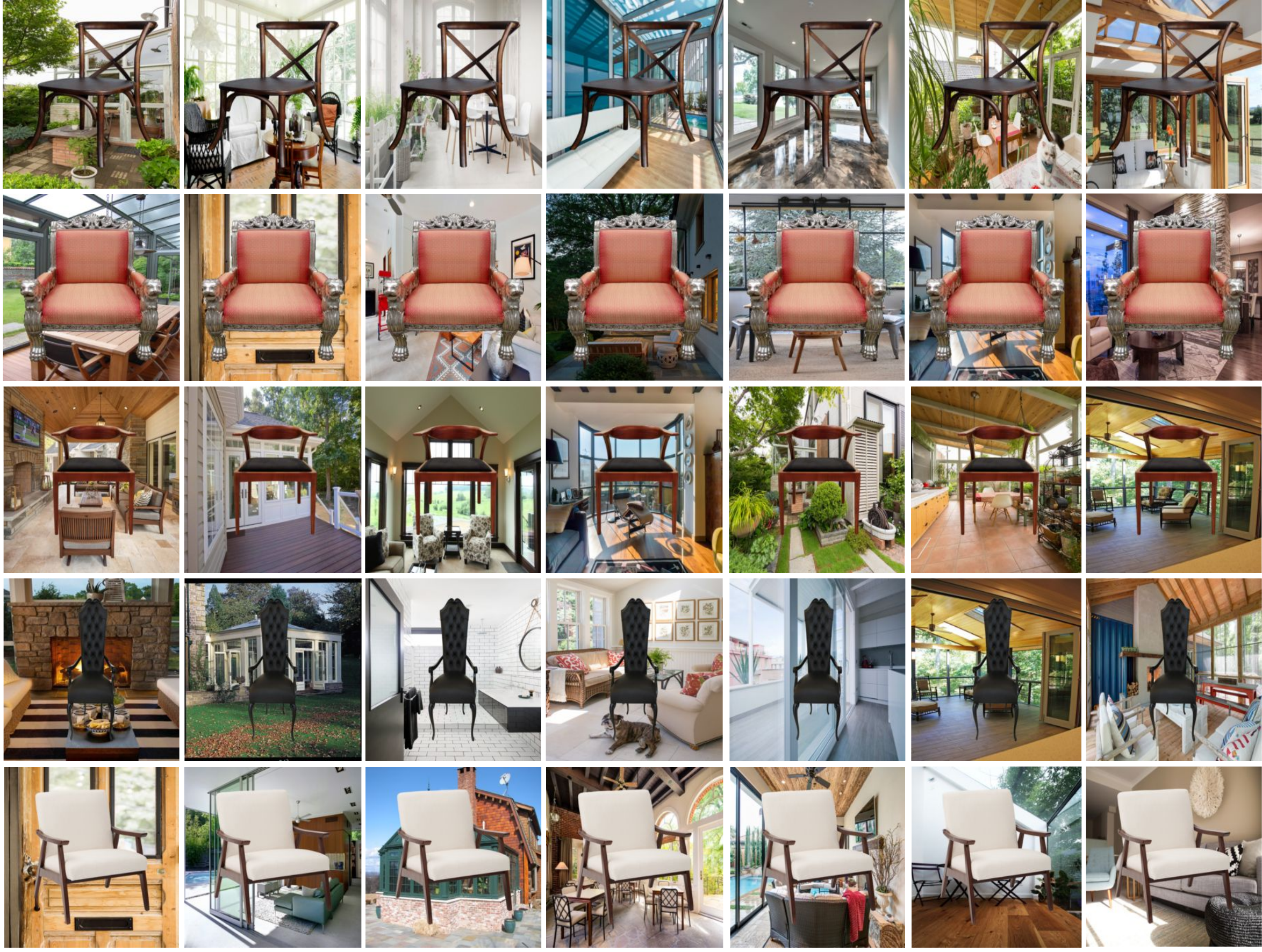}
\end{center}
   \caption{Sample images from the \ChairSegments dataset. Neither foregrounds nor backgrounds are shared across training, validation and test splits.}
   \vspace{-0.2in}
\label{fig:chair_grid}
\end{figure}

Collecting data for image segmentation is considerably costlier than for image classification. Accurate segmentation masks are laborious to annotate even for a single image. Many prior works rely on clicking to obtain object segmentations, where annotators approximate segment masks with a polygon. In our work, we rely on images with transparent backgrounds which are commonly used for e-commerce websites to display shopping items. These images are often captured using a chroma-key backdrop in order to easily and accurately separate the object from its background. By using these type of images we are able to collect thousands of object foregrounds with pixel accurate masks. 
We generate a dataset with these chair segments by compositing them on a diverse array of background images of scenes, making sure that neither segments nor backgrounds have any overlap between training, validation and testing. A sample of images in this benchmark are shown in Figure~\ref{fig:chair_grid}.

Our paper also serves to validate some of the findings of Neyshabur et al~\cite{neyshabur2020being} in transfer learning but for image segmentation. Including that models fine-tuned under the same initial pretrained model lie closer in parameter space than models trained from a random set of parameters, and that those fine-tuned models also fall within the same optimization basin compared to models trained from random initialization. As in image classification, fine-tuning also seems to bring benefits both in accuracy through feature reuse and convergence rates due to good initialization. 

Our contributions can be summarized as follows: 

\begin{compactitem}
\vspace{0.02in}
\item We introduce \ChairSegments, a semi-synthetic dataset that can be used to benchmark and quickly develop neural network architectures for object segmentation.

\vspace{0.02in}
\item Results showing that \ChairSegments is a good source of pretraining for real images by showing state-of-the-art results on the Object Discovery dataset~\cite{Rubinstein13Unsupervised}.

\vspace{0.02in}
\item Analysis of transfer learning confirming previous empirical results on transfer learning for image classification, signaling that gains in transfer learning for classification area are likely to be beneficial for image segmentation as well.

\end{compactitem}

Our paper is organized as follows: Section~\ref{sec:related} discusses related segmentation datasets and use of synthetic datasets in various applications, Section~\ref{sec:dataset} discusses the construction of Chair Segments, Section~\ref{section:analysis} states the main claims of our work with respect to our provided resource and our findings on transfer learning for segmentation, Section~\ref{section:experiments} presents our experiments on using Chair Segments and initial benchmarks, Section~\ref{section:analysistransfer} discusses our findings in transfer learning analysis, and finally Section~\ref{sec:conclusion} concludes our paper.

\section{Related Work}
\label{sec:related}

Our work is related to many previous efforts in collecting datasets for segmentation, and the use of semi-synthetic data for the development of image segmentation models.

\vspace{0.02in}
\noindent{\bf Compact Datasets} Compact datasets such as CIFAR-10 and CIFAR-100 \cite{krizhevsky2009cifar} have been heavily used to develop novel architectures for image classification. We argue that such equivalent dataset is missing for the task of image segmentation. As a result, many developments in models for segmentation have happened as separate efforts e.g. U-Net~\cite{ronneberger2015u} in the medical image domain, and fully convolutional networks (FCNs)~\cite{long2015fully} in semantic segmentation. We extensively test FCNs and U-Net models in our proposed \ChairSegments benchmark and confirm their relative strength as evidenced in larger scale benchmarks.

\begin{figure}[t!]
\begin{center}
   \includegraphics[width=0.8\linewidth]{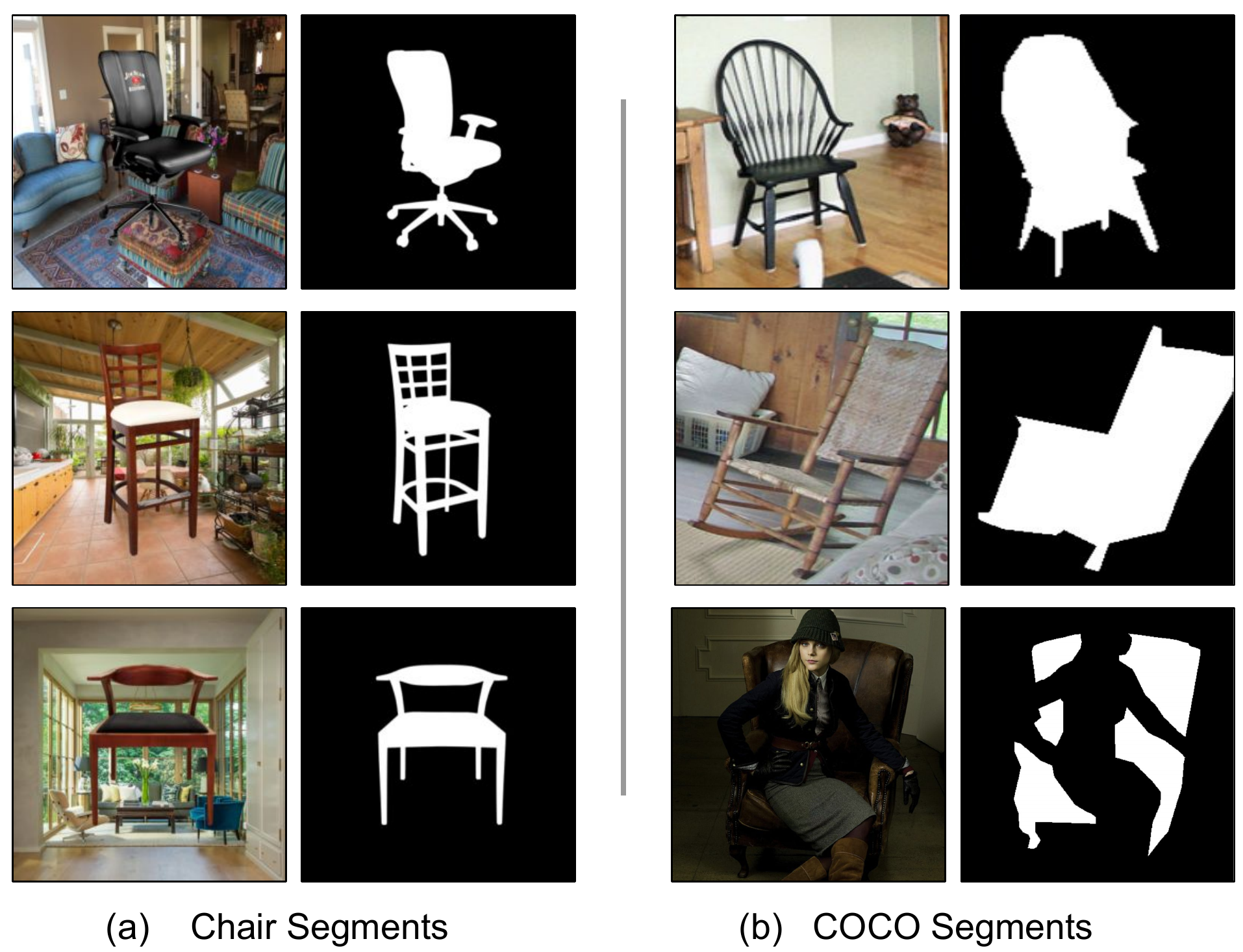}
\end{center}
\vspace{-0.1in}
   \caption{Images in the Chair Segments dataset are composites but have high resolution masks (a), while other datasets such as COCO Segments use polygon annotations which do not strictly follow object boundaries (b). Chair Segments allows designing models optimized for mask quality and are thus complementary.}
   \vspace{-0.15in}
\label{fig:chaircomparison}
\end{figure}

\vspace{0.02in}
\noindent{\bf Semantic Segmentation Datasets} There have been many datasets proposed in the past for semantic segmentation. These datasets usually extend the task of object detection to also include masks, such as the PASCAL VOC Dataset~\cite{everingham2010pascal}, the LabelMe Dataset~\cite{russell2008labelme}, and more recently the COCO Dataset~\cite{lin2014microsoft}, the KITTI dataset\cite{geiger2013vision}, and the ADE20K Dataset~\cite{zhou2017scene}, among others. Unlike these datasets \ChairSegments is built to be compact and pixel-accurate, and does not require to solve the semantic labeling task together with the pixel grouping task. This will allow researchers to concentrate in obtaining good grouping for a specific category while bypassing the need to solve together other problems such as the long-tail problem in segmentation datasets. Moreover, while Chair Segments is smaller in scale, due to its conception, it contains ground truth masks that are of considerably higher quality than existing datasets (see Figure~\ref{fig:chaircomparison}). Other compact datasets such as ObjectDiscovery~\cite{Rubinstein13Unsupervised} with excellent pixel accuracy are not large enough to allow for pretraining and explore deeply feature representation learning for segmentation. There are other datasets in specialized domains such as the INRIA Aerial Image Labeling dataset~\cite{maggiori2017dataset} in remote sensing, and the EM segmentation challenge in medical imaging~\cite{arganda2015crowdsourcing}. 

\vspace{0.02in}
\noindent{\bf Synthetic Datasets} There have been several previous efforts to use synthetic data to bypass the need to manually annotate images. For instance, dense vector field annotations for benchmarking optical flow algorithms are almost infeasible to be annotated by hand. Datasets such as the Flying Chairs dataset~\cite{dosovitskiy2015flownet} consisting of synthetic scenes, have enabled great progress in this area. For image segmentation, the SAIL-VOS Dataset used a graphics engine from a videogame to render scenes with segmentations~\cite{hu2019sail}. Unlike these datasets based on computer generated imagery, we propose composites of images with transparent backgrounds as supported by the portable network graphics format (PNG) which is widely used on the web to advertise products in image catalogs. Closest to our approach is the previous work of Jin~et~al~\cite{jin2017webly} that queried images on a search engine for various objects with white backgrounds.

\section{Data Collection: \ChairSegments}
\label{sec:dataset}

Our objective was to collect a compact dataset where model architectures can be tested in a short amount of time, while maintaining a reasonable level of difficulty. We also wanted to isolate the task of segmentation and grouping from issues of categorization. However, we do acknowledge that pixel grouping is a holistic process that requires both low-level features and top-down reasoning~\cite{arbelaez2009contours}. As a result, a generic segmentation model that is category independent is unlikely to be realized, therefore we aimed to collect a category specific segmentation resource that is instead diverse in terms of shapes and challenging in terms of visual complexity. Chairs have challenging and highly intricate shapes for image segmentation, as they often contain thin and elongated parts, hollow segments, and self-occluding parts. Moreover chairs are the hardest categories in current benchmarks such as the Pascal VOC segmentation task\footnote{Pascal VOC Segmentation task. Retrieved on November 2020. \url{http://host.robots.ox.ac.uk:8080/leaderboard/displaylb_main.php?challengeid=11&compid=6}.} where the top performer achieves an average overall precision of $90.5$ but only $57.5$ precision for the chair category.

We queried images using the Google search engine and restrict the search results to only images with transparent backgrounds. In order to ensure diversity we issue queries for chairs by relying on pairs of words from the LEVAN ontology~\cite{divvala2014learning} which was compiled in a data driven fashion. Queries included things such as ``arm chair'', ``accent chair'', ``swivel chair'', ``office chair'', ``dining chair'', etc. Then we manually filtered out images that did not contain transparent backgrounds, did not contain relevant images, or contained more than one chair in the same image. Most of the remaining chairs have masks that are of the highest accuracy as these are often used in e-commerce and are captured in controlled conditions with chroma-key backdrops and/or multiple camera settings for clean and effective background removal.
Using this process we ended up with $900$ chair segments by dropping the pixels indicated by the transparent background.  Similarly, we selected a random subset of $10,000$ background scene images following a similar protocol as in the Houzz Dataset~\cite{poursaeed2018vision} but also ensuring diversity across different types of indoor and outdoor scenes in a house environment. In this way, we encourage our foregrounds and backgrounds to be somewhat semantically consistent. Our final dataset will be released publicly along with code to replicate our current benchmarks.

We split the chair segments into $500$ for the training split, $200$ for the validation split, and $200$ for the testing split. Similarly, we split the background images into three groups containing 60\% for the training split, 20\% for the validation split and 20\% for the testing split. Then we create composites for the training set, by pairing each of the $500$ chair segments in this split with $100$ randomly selected backgrounds in the training split for every chair, resulting in a total of $50,000$ image composites for the training set. We follow a similar process to generate composite images for the validation and testing set, to obtain $20,000$ for each of them.

We generate composites at several resolutions by first resizing the background image into a target $d\times d$ resolution, and then alpha blending (with the transparent background mask) with a given chair segment that is first resized to a $(d-k)\times (d-k)$ resolution where $k$ is a padding size that is left around the chair segment. We create several versions of our dataset at various resolutions: $32\times32$, $64\times64$, $128\times128$, and $256\times256$, using padding sizes of $3$, $6$, $13$, and $26$ pixels respectively for each resolution.


\section{Main Claims}
\label{section:analysis}
In the rest of the paper we aim to demonstrate three main claims related to the proposed resource and transfer learning in image segmentation more broadly: 
\vspace{0.02in}
\begin{compactitem}
    \vspace{0.02in}
    \item \ChairSegments is challenging while allowing fast benchmarking (in a few hours) for modern neural network architectures designed for image segmentation such as U-Net~\cite{ronneberger2015u} and FCNs~\cite{long2015fully}.
    \vspace{0.02in}
    \item \ChairSegments despite being semi-synthetic, represents well the type of learning required for real datasets as evidence by state-of-the-art results on the Object Discovery Dataset~\cite{Rubinstein13Unsupervised} through transfer learning. We also compare pretraining on chair images from the existing COCO dataset vs pretraining on our proposed Chair Segments dataset with favorable results.
    \vspace{0.02in}
    \item Transfer learning in image segmentation exhibits similar properties to those recently found in the transfer learning process for image classification~\cite{neyshabur2020being}.
\end{compactitem}
\vspace{0.03in}
\begin{table*}[t]
\begin{center}
\begin{tabular}{l|cc|cc|cc|cc }
\hline
& \multicolumn{8}{c}{Resolution}\\
 & \multicolumn{2}{c}{32$\times$32}& \multicolumn{2}{c}{64$\times$64}& \multicolumn{2}{c}{128$\times$128} & \multicolumn{2}{c}{256$\times$256}\\
\hline
Model & time   & IoU & time   & IoU  & time   & IoU  & time  &  IoU \\ \hline

U-Net & 0.4 &  77.59 & 0.5 &  85.08 & 1 &  89.79 & 6.4  &91.68\\
FCN-VGG-16 & 2.1  & 60.75 & 2.7   & 61.09  & 3.6  & 74.55 & 9.1   & 68.38\\
FCN-ResNet-50 & 5.5   & 53.59 & 19.3   & 60.19 & 10.1  & 70.94  & 33   & 82.38\\

\hline
\end{tabular}
\end{center}
\caption{This table shows the time in hours  and the number of epochs that each model required to converge on \ChairSegments under various resolutions. For instance, a model such as U-Net can be trained to convergence in about $30$ minutes on images at a $64\times64$ resolution.}
\label{table:timings}
\vspace{-0.1in}
\end{table*}
The first claim is important as a trivial dataset could be potentially limited in its lifetime. However at this point it is worth noting that even datasets such as MNIST~\cite{lecun1998gradient} have withstood the test of time despite being considered largely ``solved'' -- where many modern approaches can reach accuracies close to the maximum possible. That said, we do show that Chair Segments is reasonably challenging with the top performing model obtaining an IoU of $85.08$ at our target standard resolution of $64 \times 64$ pixels. We further demonstrate the validity and usefulness of Chair Segments by showing that models that are more performant on large scale benchmarks are also more performant on Chair Segments (e.g. U-Net outperforms vanilla FCNs). 

The second claim is demonstrated by showing that by simply fine-tuning models pretrained on Chair Segments for a smaller dataset of real images. We demonstrate this by fine-tuning models on the Object Discovery Dataset~\cite{Rubinstein13Unsupervised} which contains three categories unrelated to chairs: horses, cars, and airplanes. A fine-tuned U-Net model yields state-of-the-art results on all metrics. 

The third claim is demonstrated by replicating the experiments from Neyshabur~et~al~\cite{neyshabur2020being} that demonstrate that under transfer learning, models tuned on a smaller dataset share certain properties such as fine-tuned models lying in the same optimization basin despite being trained under stochastic gradient descent (SGD). Our work is the first to empirically demonstrate these properties for transfer learning in image segmentation.

\section{Experiments}
\label{section:experiments}

We consider two approaches for image segmentation resulting in four different models. The first is the fully-convolutional network (FCN) approach~\cite{long2015fully} consisting of standard neural network architectures for classification where the last layers are replaced with an upsampling layer that produces a 2D output instead of classification scores. The second approach we consider is U-Net~\cite{ronneberger2015u}, consisting of encoder-decoder modules containing multiple downsampling and upsampling operators respectively with skip-connections to preserve high-resolution information across the network which has proved essential for segmentation. Our implementation of U-Net also relies on zero padding in order to be used with low resolution inputs. For the FCN approach we consider three variants leveraging various neural network architecture backbones, VGG-16~\cite{simonyan2014very}, ResNet-50, and ResNet-101~\cite{he2016deep}. We denote the resulting four models in our experiments as: FCN-VGG-16, FCN-ResNet-50, FCN-ResNet-101, and U-Net.

We use three standard metrics for image segmentation in our experiments:
\textbf{Precision} also known as per-pixel accuracy, which simply consists in the percentage of pixels in the image that were classified correctly; \textbf{Jaccard Index} also known as Intersection-Over-Union (IoU), which is the area of overlap between the predicted segmentation and the ground truth divided by the area of their union; and \textbf{Dice Coefficient}, also known as F1 score, which can be computed as $2\times$ the area of overlap divided by the total number of pixels between the ground truth segmentation and the segmentation predicted by the model. These are all three widely used metrics in this area.

\subsection{Training Time Trade-Offs}
First, we measure how quickly can we train to full convergence some of our models in the Chair Segments dataset and whether they are able to achieve a reasonable accuracy. We train U-Net, FCN-VGG-16 and FCN-ResNet-50 models on the \ChairSegments dataset for all resolutions and train until convergence by early stopping if the IoU metric has not changed in the past few training steps. We show results on Table~\ref{table:timings} including time to convergence and the maximum IoU score achieved.

We particularly found that U-Net can reach full convergence in four epochs in about $30$ minutes for the $64 \times 64$ resolution, and in four epochs in about $1$ hour for the $128 \times 128$ resolution. Based on these results, we decided to use the $64\times64$ resolution for the remaining experiments on the paper and this will be the recommended resolution to be used for future benchmarking. Note that this is only four times the resolution of images in the CIFAR-10 dataset that are scaled at a resolution of $32 \times 32$.

\begin{table}[b]
\begin{center}
\begin{tabular}{l|c|c|c}

\hline
Model & Prec. & IoU  & Dice \\ \hline

U-Net & 97.18 & 85.08 & 91.25 \\

FCN-VGG-16 
& 91.73 & 61.09  & 74.09\\

FCN-ResNet-50 
& 92.04 & 60.19  & 72.58\\ 

FCN-ResNet-101 
 & 92.19 & 61.62 & 73.96 \\ \hline 

\end{tabular}
\end{center}
\caption{Benchmarking results of different models on the \ChairSegments dataset. U-Net is the top performing method while fully convolutional networks (FCNs) trail in all metrics depending on the feature backbone used.}
\label{table:chairsegments}
\end{table}

\subsection{Benchmarking on \ChairSegments}
We benchmark all four models on \ChairSegments at the fixed resolution of $64 \times 64$. For FCN-VGG-16 we train with a learning rate of 1e-4, using standard stochastic gradient descent and a per pixel binary cross entropy loss. For the FCN-ResNet-50 and FCN-ResNet-101 models, we use a learning rate of 1e-5,  with an RMSprop optimizer, momentum of $0.9$, weight decay of 1e-6, and a per pixel binary cross entropy loss. For the U-Net model, we use a learning rate of 1e-3 with an Adam optimizer, and a per pixel binary cross entropy loss.

\begin{figure*}[t!]
\begin{center}
    
   \includegraphics[width=0.9\linewidth]{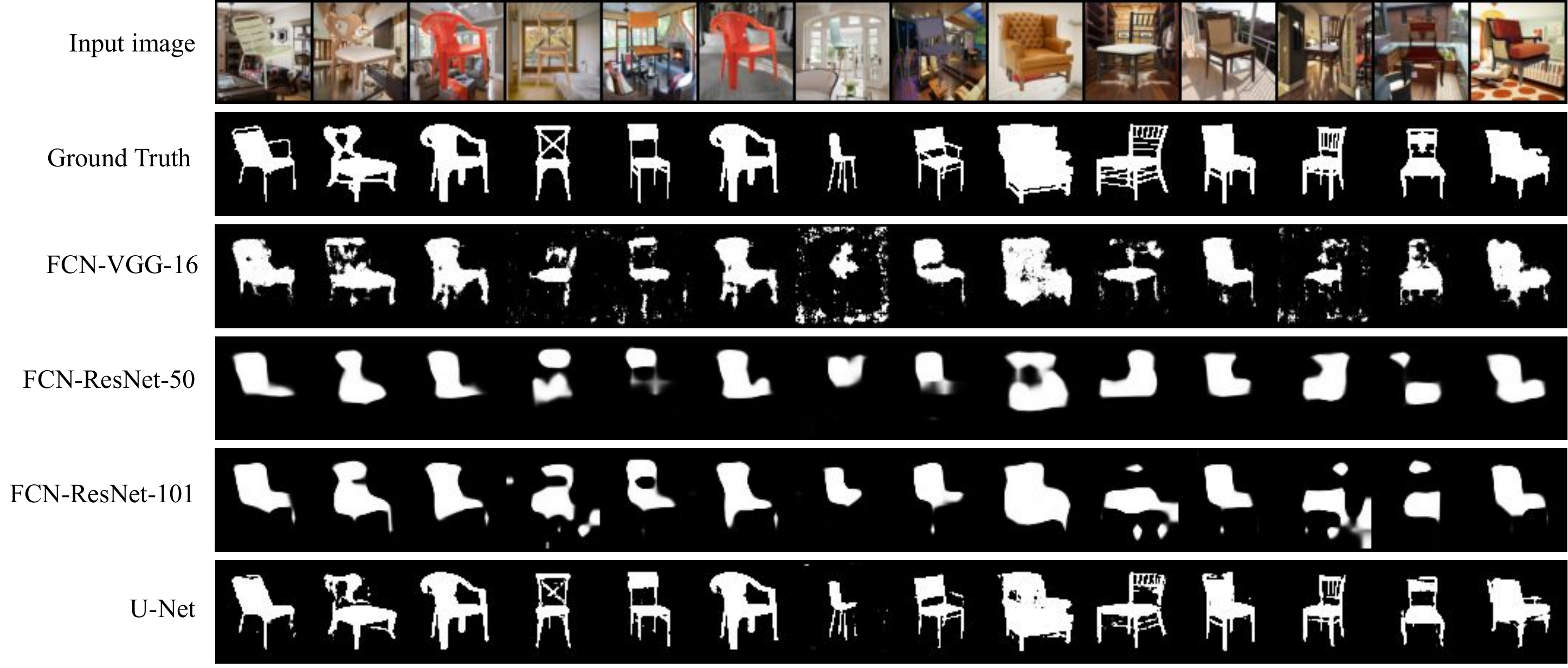}
\end{center}
   \caption{Benchmark of common generic segmentation backbones used in the literature. Results on the \ChairSegments dataset show that under several resolutions, the relative strength of these methods is preserved. We envision \ChairSegments to play the role of the CIFAR-10 dataset for quick development of algorithms for segmentation.}
\label{fig:test}
\end{figure*}

\vspace{0.02in}
\noindent \textbf{Results.} Table~\ref{table:chairsegments} shows the results for all these models across all metrics. We observe that U-Net is the top performing model by a considerable margin, followed by FCN-ResNet-101, FCN-ResNet-50 and FCN-VGG-16. Thus, demonstrating that our benchmark represents well the difficulties of larger benchmarks with respect to top performing models for image segmentation. Figure~\ref{fig:test} shows example results for these models along with the input images and ground truth segmentation masks. We observe that FCNs coupled with classification networks seem to be learning the right masks but they fail to capture finer level details as these networks were mostly designed for image classification. This is especially clear for the ResNet networks which due to their depth they inherently tend to lose high-resolution information along their computation path and multiple steps of downsampling. On the other hand, U-Net having multiple downsample and upsample steps coupled with skip connections manages to retain high-resolution information much better than the other models. This implementation of U-Net also avoids boundary issues through zero padding.

\vspace{0.04in}
\noindent \textbf{Is Chair Segments solved?} The top performing U-Net model achieves a precision of $97.18$, however a zero-knowledge baseline that blindly assigns every pixel as background would achieve a precision of $78.01$ on this dataset. The IoU metric (intersection over union) is therefore much more informative and even the best model achieves $85.08$ in this score. Models that are not able to generate precise high resolution masks lag considerably behind on this metric. We consider that models achieving IoUs squarely above $90$ will be an open area of investigation and this benchmark is unlikely to saturate soon. We expect that this leaves room for future improvements both in terms of stronger model architectures for segmentation and faster models for real time deployment.

\subsection{Fine-tuning on the Object Discovery Dataset}

\begin{table*}[h]
\begin{center}
\begin{tabular}{l|c|ccc|ccc|ccc}

\textbf{} &Pretraining &\multicolumn{3}{c}{{Car}} & \multicolumn{3}{|c|}{{Horse}} & \multicolumn{3}{c}{{Airplane}}\\

Model &  Dataset & {Prec.}  & {IoU} & {Dice} & {Prec.} & {IoU}  & {Dice}  & {Prec.} & {IoU} & {Dice}\\ \hline

FCN-VGG-16 &
--& 78.70 & 46.47 & 62.27 
& 80.54 & 36.19 & 51.57 
& 81.31 & 36.38 & 52.23\\

FCN-ResNet-50 &
--& 79.97 & 35.03  & 45.13  
& 86.46 & 26.44  & 36.79  
& 88.48 & 20.84  & 28.72\\

FCN-ResNet-101 &
-- & 94.57 & 82.73 & 88.99  
& 91.44 & 57.68 & 70.03 
& 93.12 & 60.64 & 73.08  \\ 

U-Net &
-- & 95.70 & 86.16 & 91.72 
& 93.52 & 66.41 & 74.99 
& \textbf{94.57} & 64.93 & 73.99 \\ \hline  

FCN-VGG-16  &
Chair Segments & 83.80 & 51.08 & 64.14    
& 85.69 & 33.33 & 46.37   
& 87.27 & 33.89 & 47.30\\ 

FCN-ResNet-50 &
Chair Segments& 85.59 & 53.61  & 63.44   
& 89.26 & 41.08 & 53.61         
& 90.79 & 38.82 & 50.93\\ 

FCN-ResNet-101 &
Chair Segments & 93.05 & 81.34 & 87.77         
& 90.38 & 57.73 & 70.28           
& 91.60 & 51.54 & 51.54\\

\hline

U-Net &
COCO Chairs & 95.96 & 87.17 & 91.51
& 94.20 & 68.61 & 77.43 
& 94.19 & 68.05 & 77.53\\

U-Net &
Chair Segments& \textbf{96.43} & \textbf{88.22} & \textbf{92.23}
& \textbf{94.46} & \textbf{70.47} & \textbf{78.47}
& 94.49 & \textbf{71.59} & \textbf{80.42}\\ 

\end{tabular}
\end{center}
\caption{Pretraining on \ChairSegments demonstrates a clear advantage overall and almost across all metrics in all subsets.}
\label{table:obj-discovery-results}
\end{table*}

\begin{table*}[h]
\begin{center}
\begin{tabular}{l|cc|cc|cc}
\hline
\textbf{} & \multicolumn{2}{|c|}{{Car}} & \multicolumn{2}{c|}{{Horse}} & \multicolumn{2}{c}{{Airplane}}\\

Model & Prec.  & IoU & Prec. & IoU   & Prec. & IoU \\ \hline

Rubinstein et al.\cite{Rubinstein13Unsupervised}
& 85.4 & 0.64
& 82.8 & 0.52
& 88.0 & 0.56 \\

Quan et al.\cite{Quan_2016_CVPR}
& 88.5 & 0.67
& 89.3 & 0.58
& 91.0 & 0.56  \\

Jerripothula et al.\cite{jerripothula2016image}
& 88.0 & 0.71            
& 88.3 & 0.61              
& 90.5 & 0.61  \\

Yuan et al.\cite{yuan2017deep} 
& 90.4 & 0.72                
& 90.2 & 0.65               
& 92.6  & 0.66 \\

U-Net 
& 95.7 & 0.86
& 93.5 & 0.66
& \textbf{94.6} & 0.65\\

U-Net (w/ pretraining)
& \textbf{96.4} & \textbf{0.88} 
& \textbf{94.5} & \textbf{0.71}
& 94.5 & \textbf{0.72} \\ \hline

\end{tabular}
\end{center}
\caption{U-Net pretrained on \ChairSegments obtains the state-of-the-art in the Object Discovery dataset. }
\label{table:object-discovery-state-of-the-art}
\end{table*}

The Object Discovery dataset introduced by Rubinstein~et~al~\cite{Rubinstein13Unsupervised} for object segmentation contains images with ground truth masks for three types of objects: cars, horses and airplanes. This dataset although small, has been used over the years to test from handcrafted methods based on energy-based minimization and graphical models, to more recent neural network architectures for segmentation. We train our four considered models on this dataset in two conditions: from randomly initialized weights, and from weights pretrained on \ChairSegments. Additionally, we train the best model (U-Net) with chair segments obtained from the COCO Segments dataset (COCO Chairs) instead of our dataset. Our goal is to measure whether \ChairSegments provides useful transfer learning information for this task.

For FCN-VGG-16 we train with a learning rate of 1e-4, an SGD optimizer with per pixel binary cross entropy loss. 
For the FCN-ResNet-50 model we use a learning rate of 1e-4, an SGD optimizer with momentum of 0.9, weight decay of 1e-5, and a per pixel binary cross entropy loss. 
For the FCN-ResNet-101 model we use a learning rate of 1e-5, an RMSprop optimizer with a momentum of $0.9$, weight decay of 1e-6, and a per pixel binary cross entropy loss. 
Finally, for U-Net we use a learning rate of 1e-3, and an Adam optimizer with a per pixel binary cross entropy loss. These hyperparameters were determined using the validation split through standard hyperparameter tuning. 

\vspace{0.04in}
\noindent \textbf{Results and Discussion.} Table~\ref{table:obj-discovery-results} shows results for all the models both when training from scratch (from randomly initialized weights), and when fine-tuning from a model pretrained on the \ChairSegments dataset. We additionally show results for U-Net when pretrained on COCO Chairs. We observe as in the \ChairSegments dataset, that U-Net is the superior model across all experiments, and fine-tuning helps in almost all subsets of this dataset under all metrics, and often by a considerable margin especially on the IoU metric which is the most important for segmentation quality. Moreover, U-Net pretrained on the \ChairSegments dataset outperforms U-Net pretrained on COCO Chairs, further demonstrating that \ChairSegments is a valuable resource and is able to represent well a segmentation task with real-world images despite being semi-synthetic. 

\begin{figure*}[t]
\begin{center}
   \includegraphics[width=0.88\linewidth]{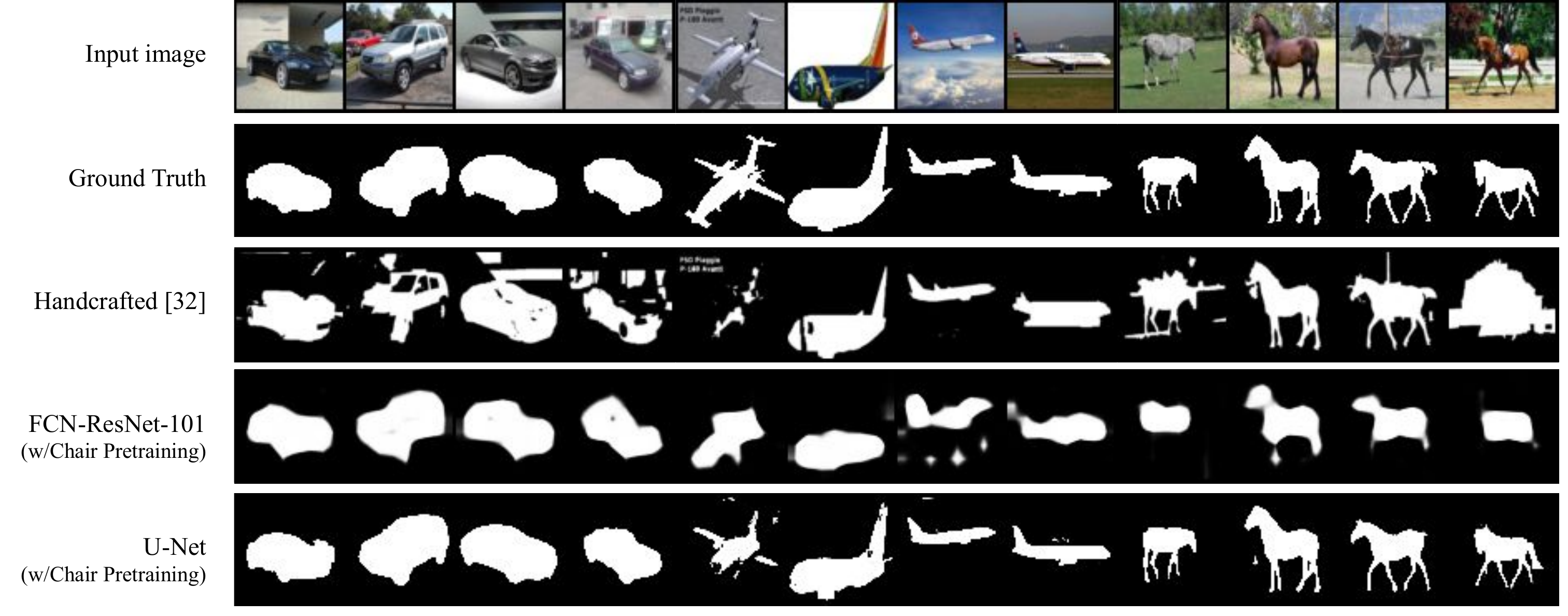}
\end{center}
   \caption{Qualitative results for segmentation models on the Object Discovery dataset, along with their ground truth segmentation masks.}
   \vspace{-0.1in}
\label{fig:test2}
\end{figure*}

Table~\ref{table:object-discovery-state-of-the-art} shows our top performing models U-Net and U-Net pretrained on \ChairSegments when compared against all previous methods in the research literature, establishing these two models as the state-of-the-art under this benchmark. Figure~\ref{fig:test2} shows qualitative results on this benchmark for a few examples along with results released with this dataset~\cite{Rubinstein13Unsupervised} using a hand-crafted approach. We can see that FCN-ResNet-101 still struggles with finer-level details while U-Net is able to produce masks at a much higher resolution. However, even U-Net seems to be assigning some pixels as foreground when they are clearly disconnected or dissimilar from the object, which leaves room for future improvement in model designs that can better preserve finer-level structures.

\section{Transfer Learning Analysis}
\label{section:analysistransfer}
There has been recent work in image classification to analyze the extent to which transfer learning helps in tasks that have access to less amounts of training data~\cite{ngiam2018domain,he2019rethinking,kornblith2019better,neyshabur2020being}. More specifically, there is a growing interest to analyze to what extent does transfer learning help with providing a better starting point for optimization and to what extent feature reuse is the driving factor. We adopt the recently proposed framework of Neyshabur~et~al~\cite{neyshabur2020being} which allows us to directly draw analogies between transfer learning in image classification and transfer learning in segmentation.

\begin{figure}[t!]
\begin{center}
   \includegraphics[width=0.74\linewidth]{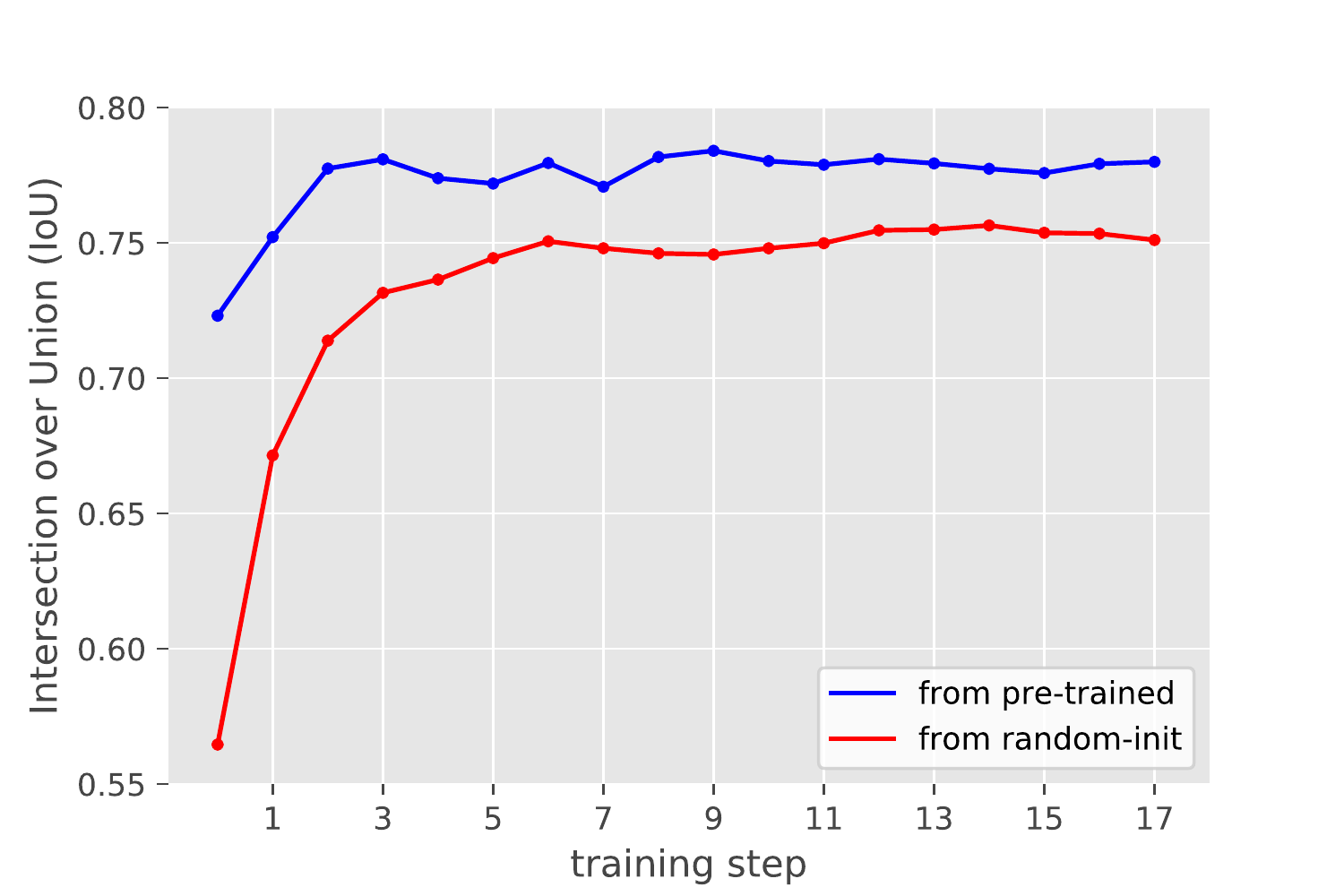}
\end{center}
\vspace{-0.1in}
   \caption{Fine-tuning from a pretrained model on \ChairSegments vs training from scratch using randomly initialized weights averaged across epochs and two runs. \ChairSegments demonstrates gains on a real dataset --ObjectDiscovery-- despite it being a semi-synthetic dataset.}
\label{fig:finetuning-test}
\end{figure}

We show in Figure~\ref{fig:finetuning-test} a plot averaging two runs of our U-Net model training from scratch, and fine-tuned from pretrained weights across training steps until convergence. We can observe that not only the pretrained model is reaching a higher IoU score but it also reaches a value close to convergence much faster than the same network trained from random parameters. Hence we can see that both feature reuse and good initialization are driving factors -- confirming results obtained for image classification. This result might seem trivial but we would like to emphasize the discrepancy in the source and target domains, where the source domain only contains chair images and chair segments, while the target domain only contains, horses, cars and airplanes.

There are two additional findings that we present in this section: (1) Analyzing how far apart are comparable pretrained models and models trained from scratch in terms of $\ell_2$ distances, and (2) Exploring the linear interpolation space between models trained from pretrained weights, and models trained from random initialization. In both cases, we found that the results we obtain confirm those obtained for image classification in Neyshabur~et~al~\cite{neyshabur2020being}. These results could lead to better algorithms for initializing neural networks for segmentation and for optimizing ensembles of models that lie within close areas in the optimization landscape.

\begin{table}[h]
\begin{center}
\setlength{\tabcolsep}{4pt}
\begin{tabular}{c|c|c|c|c}
\hline
Model & Ft \& Ft & Ri \& Ri & Ft \& P & Ri \& P \\ \hline

Classification~\cite{neyshabur2020being} & 193.45 & 815.08 & 164.83 & 796.80\\ 

Segmentation (ours) 
& 373.94 & 637.05  & 313.06 & 549.94\\ \hline

\end{tabular}
\end{center}
\caption{We examine the $\ell_2$ distance between a pretrained model (P), fine-tuned models (Ft), and models trained from random initialization (Ri). Following~\cite{neyshabur2020being}, we find that two models trained from scratch lie further apart than two models fine-tuned from a common pretrained model, confirming their findings but for image segmentation. In this experiment, model P is pretrained on \ChairSegments and fine-tuned on ObjectDiscovery. }
\label{table:distances}
\vspace{-0.1in}
\end{table}

\subsection{Analyzing Model Distances in Parameter Space}
Table~\ref{table:distances} shows $\ell_2$ distances in parameter space between two models fine-tuned from the same set of pretrained weights (Ft \& Ft) for both image classification and image segmentation (first column), similarly we compute $\ell_2$ distances between two models trained from randomly initialized weights (Ri \& Ri) for both tasks (second column), and between a model fine-tuned from pretrained weights and the pretrained network itself (Ft \& P) for both tasks (column 3), and between a network trained from randomly initialized weights and a pretrained network (Ri \& P) (last column). For all segmentation models we used the U-Net neural network architecture.

The results are strikingly similar as those obtained for image classification by~\cite{neyshabur2020being}, suggesting that improved methods for transfer learning in image classification are likely to lead to gains in image segmentation as well. And more importantly for us, that our fine-tuned models behave similarly as models fine-tuned from Imagenet pretrained models for image classification on a smaller dataset -- which suggests that \ChairSegments is useful as a proxy for real world data. To summarize these results, it seems models that are trained from the same initialization from pretrained weights end up lying closer in parameters space, than models that started from random weights instead. As a sanity check, unsurprisingly, models that are fintuned, are closer in parameter space from their starting point pretrained network (Ft \& P) compared to models trained from scratch (Ri \& P).

\subsection{Analyzing the Loss Landscape of Models}
Finally, we analyze the behavior of our pair of models (Ft \& Ft), both trained from the same checkpoint $P$, we shall denote them more formally here $F_1$ and $F_2$, and the two models trained from randomly initialized weights (Ri \& Ri), we shall denote them more formally here as $S_1$ and $S_2$. As in the previous section, for all models we used the U-Net neural network architecture. 

\begin{figure}[h!]
\begin{center}
\vspace{-0.1in}
   \includegraphics[width=0.7\linewidth]{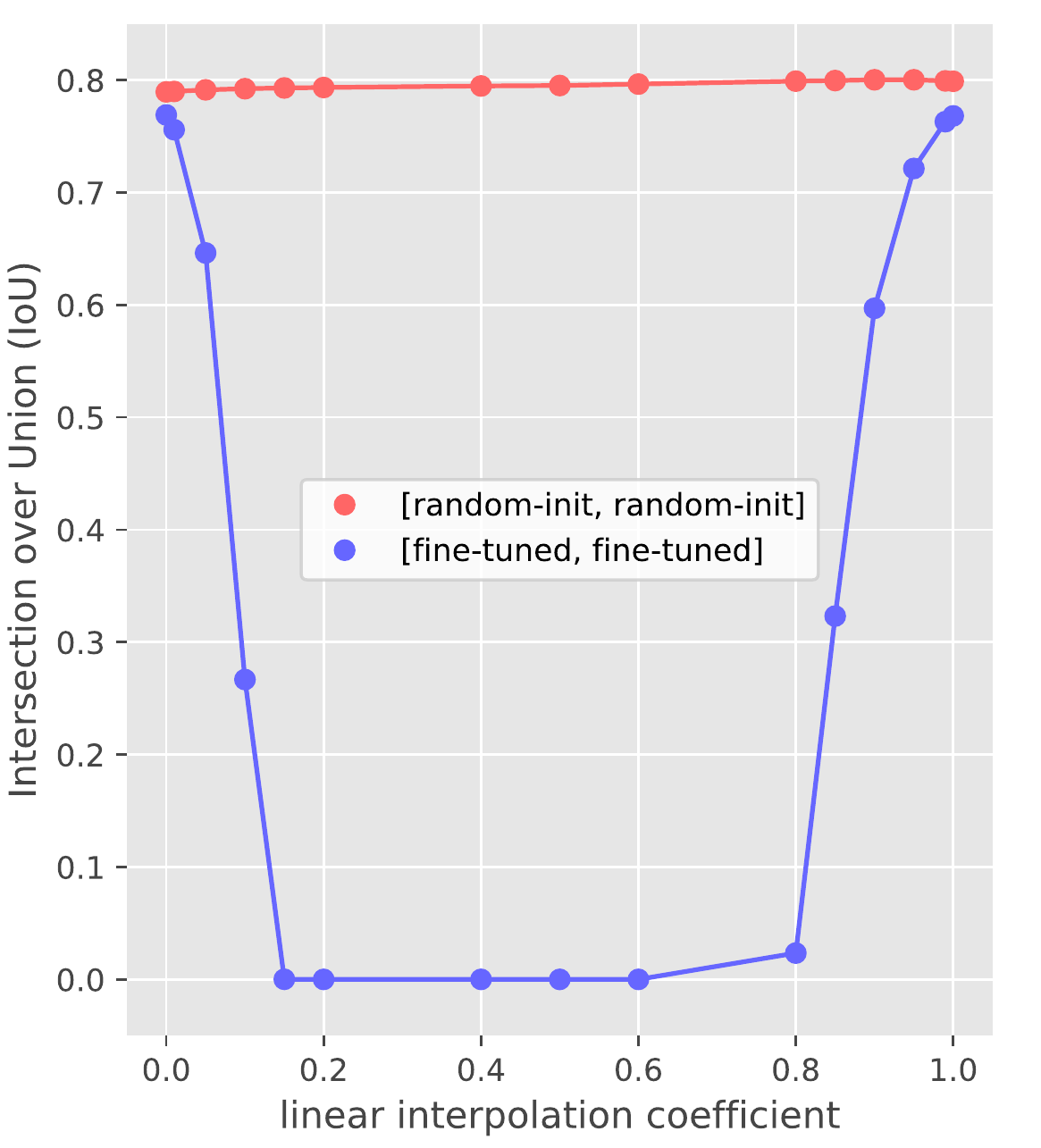}
\end{center}
\vspace{-0.1in}
   \caption{We show that two models that are fine-tuned from the same pretrained model lie within the same optimization basin with respect to a linear interpolation, while the same is not true for two models trained from scratch.}
   \vspace{-0.1in}
\label{fig:testxx}
\end{figure}

Let $W_1$ and $W_2$ be two different set of weights for models $F_1$ and $F_2$, then we can define a new model $F_\alpha$ that is a model with a set of parameters $W_\alpha = \alpha W_1 + (1 - \alpha) W_2$, where $\alpha$ is an interpolation scalar parameter betwen 0 and 1. Therefore $F_\alpha$ is a model that lies in the linear interpolation space between $F_1$ and $F_2$. We define in the same way a model $S_\alpha$ with respect to $S_1$ and $S_2$. The question we are trying to assess is: Are models that lie in between these two models found through stochastic gradient descent any good? If the loss landscape where the two models ($S_1$, $S_2$) or ($F_1$, $F_2$) lie within a flat landscape, then the transition must be smooth and intermediate models should be performant. We found this to be the case only for the pair of models $(F_1, F_2)$ as evidence in our Figure~\ref{fig:testxx} where we can clearly observe that randomly initialized models quickly degrade when moving away from the starting points, while fine-tuned models transition smoothly without loss in IoU scores. These results confirm the results in transfer learning obtained for Imagenet pretrained networks on transfer learning as evidenced in~\cite{neyshabur2020being}.

\section{Conclusions}
\label{sec:conclusion}

We can conclude that our \ChairSegments benchmark is a useful dataset that presents consistent results over different image segmentation methods and can be a useful resource to develop new models and algorithms for image segmentation that can later be applied to semantic image segmentation or semantic instance segmentation. Also, our study of transfer learning over image segmentation give us the opportunity for a better understanding of how deep neural networks behave when fine-tuned for a target task with more constrained numbers of training samples -- mirroring results observed in the more well studied problem of image classification. As a result of our experiments we are planning to apply variations of our current U-Net model to either obtain more performant models or devising new methods for optimizing models that take better advantage of the properties of transfer learning. 

\paragraph{Acknowledgments} We thank the UVA Department of Computer Science in Partnership with DREU CRA-W, for hosting and funding I.K.T and R.G. through NSF \#CNS-1246649, IAAMCS. This work was also partially funded with generous gifts from Leidos and Adobe.

{\small
\bibliographystyle{ieee_fullname}
\bibliography{egbib}
}

\end{document}